\documentclass[letterpaper, 10 pt, conference]{ieeeconf}  
\usepackage[colorlinks=true, linkcolor=blue, citecolor=blue, urlcolor=blue]{hyperref}
\usepackage{graphicx}
\usepackage{tabularx}

\usepackage{xcolor}
\usepackage{soul}
\sethlcolor{yellow}

\usepackage{pifont} 
\usepackage{booktabs} 
\usepackage{array}    
\newcommand{\cmark}{\ding{51}}  
\newcommand{\xmark}{\ding{55}}  

\IEEEoverridecommandlockouts                              

\title{\LARGE \bf
Guide-LLM: An Embodied LLM Agent and Text-Based Topological Map for Robotic Guidance of People with Visual Impairments}

\author{Sangmim Song$^{1}$, Sarath Kodagoda$^{1}$, Amal Gunatilake$^{1}$, Marc G. Carmichael$^{1}$  \\ Karthick Thiyagarajan$^{2}$ and Jodi Martin$^{3}$  
\thanks{This research was supported by the Australian Government through the Australian Research Council's Linkage Projects funding scheme (LP220100430) and the industry partner Guide Dogs NSW/ACT.}
\thanks{$^{1}$Sangmim Song, Sarath Kodagoda, Amal Gunatilake and Marc G. Carmichael are with the Robotics Institute, Faculty of Engineering and Information Technology, University of Technology Sydney, Broadway, Ultimo NSW 2007, Australia.
Email: \tt{\small{Sangmim.Song@student.uts.edu.au}}
}%
\thanks{$^{2}$Karthick Thiyagarajan is with the Smart Sensing and Robotics Laboratory (SensR Lab), Centre for Advanced Manufacturing Technology (CfAMT), School of Engineering, Design and Built Environment (SoEDBE), Kingswood, NSW 2747, Australia.
}%
\thanks{$^{3}$Jodi Martin is with Guide Dogs New South Wales /
Australian Capital Territory, Sydney, New
South Wales, Australia.
}
}
\begin{document}
\maketitle
\thispagestyle{empty}
\pagestyle{empty}


\begin{abstract}

Navigation presents a significant challenge for persons with visual impairments (PVI). While traditional aids such as white canes and guide dogs are invaluable, they fall short in delivering detailed spatial information and precise guidance to desired locations. Recent developments in large language models (LLMs) and vision-language models (VLMs) offer new avenues for enhancing assistive navigation. In this paper, we introduce Guide-LLM, an embodied LLM-based agent designed to assist PVI in navigating large indoor environments. Our approach features a novel text-based topological map that enables the LLM to plan global paths using a simplified environmental representation, focusing on straight paths and right-angle turns to facilitate navigation. Additionally, we utilize the LLM's commonsense reasoning for hazard detection and personalized path planning based on user preferences. Simulated experiments demonstrate the system's efficacy in navigating indoor environments based on user queries, underscoring its potential as a significant advancement in assistive technology. The results highlight Guide-LLM's ability to offer efficient, adaptive, and personalized navigation assistance, pointing to promising advancements in this field.
\end{abstract}
{\fontsize{8}{8}\selectfont \textbf{Code and replication details:} \url{https://github.com/awd1779/Guide-LLM}} \newline

\section{Introduction}

Navigating everyday environments can be particularly challenging for persons with visual impairments (PVI), who often depend on specialized tools, help from others, or familiar routes to get around \cite{intro5,intro1,intro4,intro2,intro3}. Traditional aids, such as white canes and guide dogs, are essential parts of the navigation; however, with the technological developments, further assistance could be possible to improve user confidence in navigation \cite{giudice2020use}. 

Recent breakthroughs in artificial intelligence, especially in large language models (LLMs) \cite{gpt,llama} and vision-language models (VLMs) \cite{llava, clip, blip}, have created new opportunities in human-robot interaction \cite{LLMHRI}, task planning \cite{sayplan, Self-planning}, and navigation \cite{LMNAV, DriveLLM}. Despite these advancements, their application in assisting PVI with navigation is still largely unexplored. \cite{dragon} demonstrated the potential of using a robot platform combined with a language model to assist PVI, offering a glimpse of what is achievable. However, fully harnessing the capabilities of LLMs and VLMs for guiding PVI remains largely unexplored.

 Conventional navigation systems typically depend on pre-programmed rules and sensor data, which may overlook the subtleties and complexities of real-world environments. In contrast, LLMs can analyze contextual information and anticipate potential hazards, offering a more adaptive and responsive solution for navigation. A key challenge in creating LLM and VLM-based navigation systems for PVI is their dependence on precise, explicit commands from users, which can be difficult for PVI. While 3D reconstruction techniques using point clouds \cite{VLMAPS, ll3da, lidarllm} and traditional methods like SLAM (Simultaneous Localization and Mapping) help LLMs understand the environment, their scalability is constrained by the high computational demands of having LLMs interpret these dense maps \cite{3d-llm,pointllm,lidarllm}.

\begin{figure}[!t]
  \centering
  \includegraphics[width=\columnwidth]{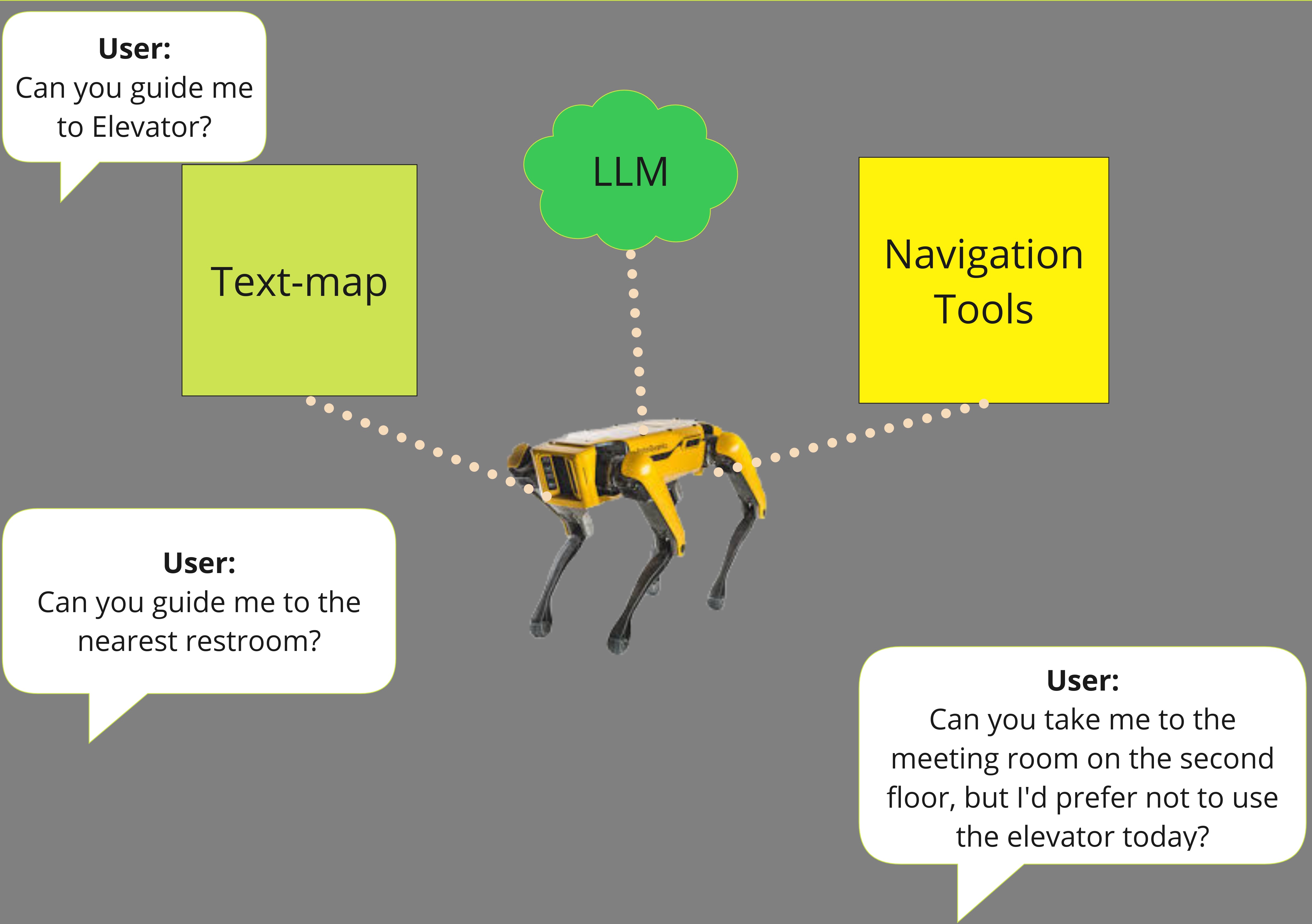}
  \caption{Guide-LLM: The embodied agent consists of a text map, LLM, and navigational modules to guide the user to the destination.}
  \label{fig1}
\end{figure}

\begin{figure*}[!t]
  \centering
  \includegraphics[width=0.9\textwidth,clip,trim=0 0 0 0]{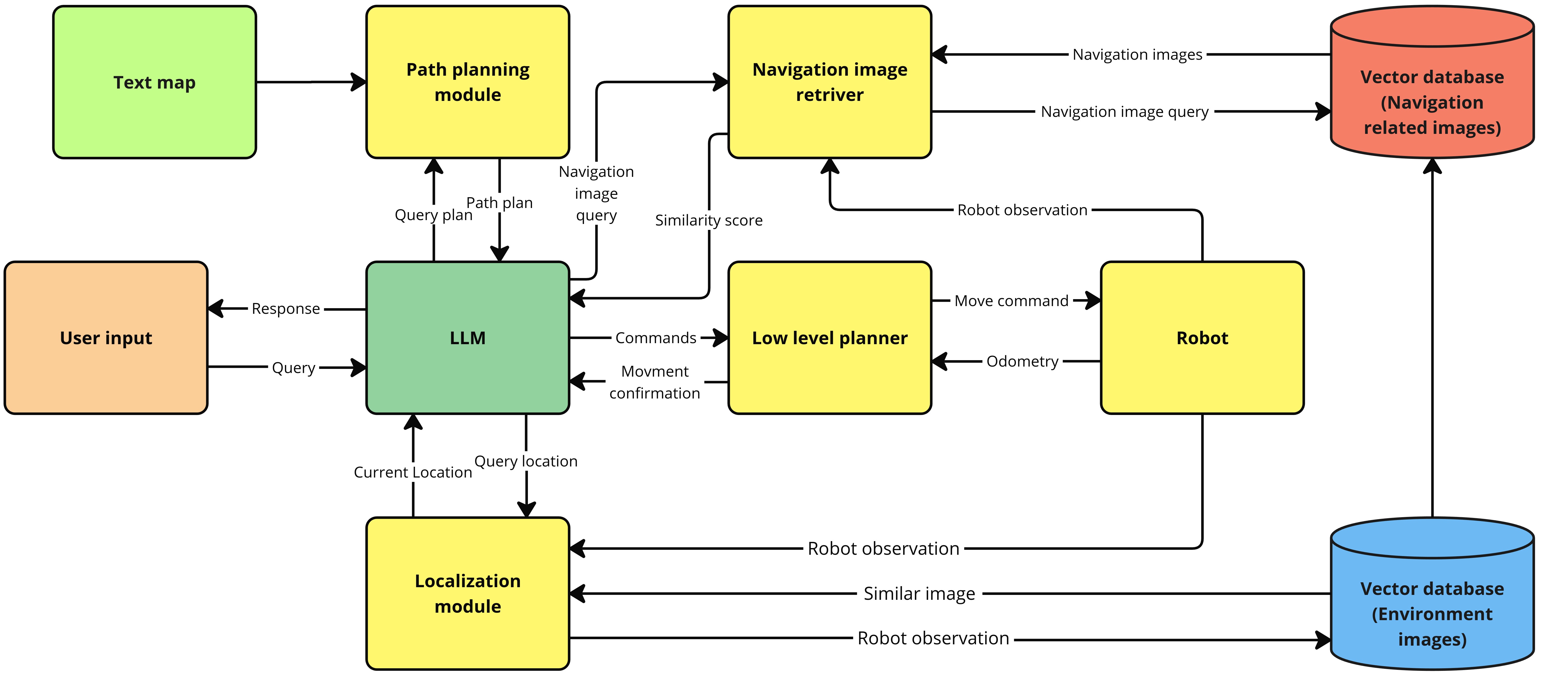}
  \caption{
Guide-LLM framework: LLM (green) serves as the central controller, using commonsense reasoning to interpret user queries and interact with various modules (yellow) for decision-making and navigation tasks. Text map (Green) provides a textual representation of the environment used by the path planning module to create route plans. Vector database 1 (blue) stores static embeddings of the environment images, aiding in consistent localization. Vector database 2 (red) stores navigational image embedding that can be updated or deleted based on the agent's requirements. }
  \label{frame-work}
\end{figure*}

To overcome this challenge, we propose an innovative framework that utilizes a text-based topological map of the environment. This allows the LLM to plan global paths by referring to a textual representation, eliminating the need for explicit user input. This approach is more computationally efficient and scalable compared to methods that rely on dense maps or 3D representations for each user query, which could cause delays, leaving PVI waiting for the LLM to process these complex inputs. Additionally, our text-based topological map is designed to address the specific needs of PVI by generating straight paths and right-angle turns, which are easier to navigate and help maintain spatial orientation \cite{swobodzinski2009indoor,shafique2024path}. These clear and predictable routes reduce cognitive effort, enabling more efficient and safer navigation, especially when compared to the challenges posed by curved or irregular pathways.

Guide-LLM also incorporates an image retrieval system for localization and a low-level planner to handle robot movement, constraining the robot’s actions to predictable patterns. One of the significant advantages of integrating LLMs into navigation is their ability to leverage commonsense reasoning, which has the potential for enhancing safety, personalized navigation, and the interpretability of its actions \cite{DriveLLM,zhou2024navgpt}, which traditional navigation systems often lack. The main contributions of our work are:

\begin{itemize}
  \item We propose Guide-LLM, an innovative framework that employs LLM as embodied agents to assist visually impaired individuals in navigation.
  \item By integrating a novel text-based topological map, LLM, and navigational tools (Fig. \ref{fig1}), our approach enables high-level planning while minimizing the need for users to provide extensive descriptions of destination.
  \item In comprehensive simulations, we validate our method’s localization error detection and recovery, personalization potential, and hazard detection demonstrating the robust capabilities required for guiding PVI.
\end{itemize}

\begin{figure*}[!t]
  \centering
  \includegraphics[width=0.9\textwidth,clip,trim=0 0 0 0]{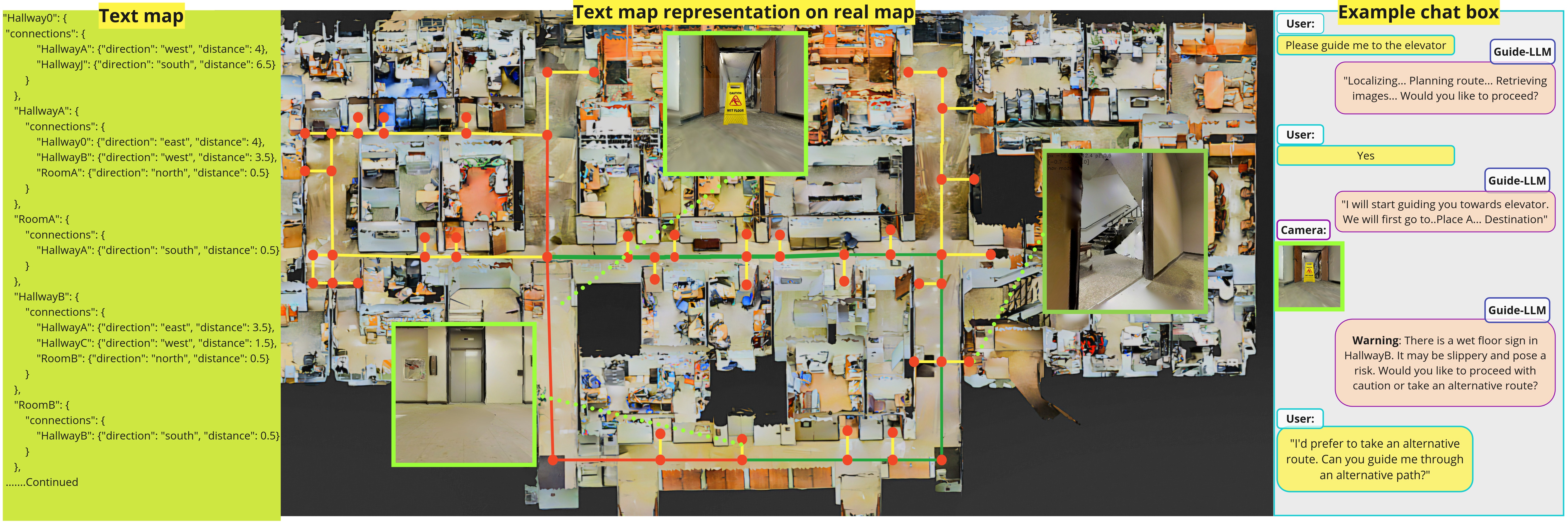}
  \caption{
Text map (Left): diagram of text map, part of the text map is extracted. Example text map representation (Middle): User asks the agent to navigate to the elevator. Guide-LLM plans a route (red line) and begins guiding. Along the route, a hazard is detected (wet floor sign), Guide-LLM warns the user and suggests an alternative path (green line), and the chat box (right) shows an example communication between Guide-LLM and the user. }
  \label{textmap+diagram}
\end{figure*}

\section{Related Work}

\subsection{Navigation Aid for People with Visual Impairment}
Recent advancements in assistive technology have greatly enhanced the navigation and mobility of PVI. For instance, the robotic white cane \cite{whitecane} employs multimodal sensing and steering assistance to aid users in navigating their surroundings. Additionally, wearable systems have been developed to improve situational awareness by delivering real-time feedback through auditory or haptic signals \cite{wearable1,wearable2,wearable3,wearable4}. \cite{waerable5} introduced a method that allows PVIs to jog on athletic tracks. Despite these advancements, many of these systems are pre-programmed with fixed rules and behaviors, which can limit their adaptability to dynamic scenarios.
To address these limitations, dialogue-based robots have been proposed as an alternative form of assistance, using conversational interfaces to help users reach their destinations \cite{dragon}. However, these systems still encounter significant challenges, particularly in natural language interaction and adaptability. Their rule-based nature also limits their ability to generalize and adjust to changing environments, which is essential for offering personalized and flexible support to PVI users.

\subsection{Large Language Models in Robotics}
The integration of LLMs into robotics has demonstrated significant potential in areas such as task planning, autonomous driving, multimodal reasoning, and navigation. Frameworks like SayPlan and ReAct enhance robotic capabilities by breaking down complex instructions into actionable sub-tasks and combining reasoning with action \cite{sayplan,react}. DriveLLM \cite{DriveLLM} showcased how LLMs can improve interpretability and decision-making in autonomous driving by integrating object-level vector modalities with LLMs, thus enhancing context understanding and explainability in driving scenarios. RoboVQ \cite{sermanet2024robovqa} combined LLMs with vision inputs, enabling robots to perform complex, long-horizon tasks and demonstrating LLMs' multimodal reasoning capabilities. \cite{shah} utilized LLMs as navigation agents, leveraging their reasoning capabilities to develop search heuristics for exploring new environments. SayNav employs LLMs to guide robots through unfamiliar environments by grounding high-level instructions in spatial contexts, highlighting the potential of LLMs in exploration without requiring detailed prior knowledge \cite{rajvanshi2024saynav}. \cite{huang2022language} demonstrated LLMs' ability to generalize across tasks, allowing robots to plan and execute complex actions with minimal task-specific training. Additionally, NavGPT \cite{zhou2024navgpt}, MAP-GPT \cite{chen2024mapgpt}, and LGX \cite{CatshapeMug} explore LLM-guided robotic navigation, further extending LLM capabilities in uncharted environments.
Despite these advancements, there is limited research specifically focusing on using LLMs' commonsense reasoning and contextual understanding to assist people with visual impairments. This gap highlights the need for innovative approaches that harness LLMs’ reasoning abilities to provide more intelligent and adaptive support for PVI.

\subsection{Vision and Language Navigation}
Vision and language navigation (VLN) integrates visual perception with natural language understanding to enable agents to navigate based on spoken or written instructions. A significant challenge in VLN is that natural language instructions often emphasize high-level decisions and landmarks, frequently lacking detailed, low-level movement guidance \cite{fried2018speaker}. Attention-based mechanisms \cite{ma2019self,li2019vision,vasudevan2021talk2nav} and reinforcement learning approaches \cite{wang2018look,wang2020soft} have shown promising results in addressing this issue. Recently, the rise of LLMs and VLMs has sparked interest in leveraging these models to enhance VLN capabilities. For instance, LM-Nav \cite{LMNAV} uses LLMs to extract landmarks from user queries and VLMs to ground these landmarks in the environment for navigation. Similarly,  \cite{garg2024robohop} employs LLMs to translate user queries into actionable navigation tasks and use image segmentation techniques to create a topological map for navigation. \cite{CatshapeMug} combined LLMs commonsense reasoning with VLMs to navigate towards uniquely described objects, showcasing LLMs potential in understanding nuanced language. \cite{schumann2024velma} utilized LLMs to guide the agent by processing action sequences based on text prompts that include navigation instructions, visual landmark descriptions, and the agent's past trajectory. \cite{dragon} used LLMs to identify landmarks, describe the environment, and navigate through it.
Despite these advancements, navigation for PVI remains challenging, particularly due to difficulties in describing specific scenes and landmarks, which can impede effective communication with the agent. Our work aims to address this limitation by enabling our agent to infer user intent and automatically resolve ambiguities in their instructions.

\setlength{\tabcolsep}{3pt}
\begin{table*}[!t]
  \centering
  \caption{Feature Comparison Table of Our Method and Others}
  \label{tab:improved_comparison}
  \begin{tabular}{
    >{\raggedright\arraybackslash}p{0.16\linewidth}%
    >{\centering\arraybackslash}p{0.16\linewidth}%
    >{\centering\arraybackslash}p{0.14\linewidth}%
    >{\centering\arraybackslash}p{0.16\linewidth}%
    >{\centering\arraybackslash}p{0.16\linewidth}%
    >{\centering\arraybackslash}p{0.14\linewidth}}
    \toprule
    \textbf{Approach} &
    \textbf{Natural language interactions} &
    \textbf{Doesn't Require Detailed Goal Knowledge} &
    \textbf{Localization Error \& Recovery} &
    \textbf{Identify potential Hazards} &
    \textbf{Personalization} \\ 
    \midrule
    \textbf{Guide-LLM} \newline (Our Approach) &
    \cmark  &
    \cmark  &
    \cmark  &
    \cmark  &
    \cmark  \\
    \midrule
    \textbf{Traditional Navigation Aid} \newline \cite{whitecane}\,\cite{Jawale2017} &
    \xmark &
    \xmark &
    \xmark &
    \cmark &
    \xmark \\
    \midrule
    \textbf{Wearable System} \newline \cite{wearable1,wearable2,wearable3,wearable4} &
    \xmark &
    \xmark &
    \xmark &
    \cmark &
    \xmark \\
    \midrule
    \textbf{VLN Approach} \newline \cite{LMNAV}\,\cite{dragon}\,\cite{CatshapeMug}\,\cite{garg2024robohop} &
    \cmark &
    \xmark &
    \xmark &
    \xmark &
    \xmark \\
    \bottomrule
  \end{tabular}

\end{table*}


\section{Proposed System}
In this section, we provide a detailed overview of the Guide-LLM framework. Figure \ref{frame-work} illustrates the primary components of the framework, which enable our LLM agent to handle decision-making, high-level planning, and navigation.

\subsection{Core Components} 
\subsubsection{Agent} We used GPT-4o \cite{openai_gpt4} as the central component of our framework. As shown in Fig. \ref{frame-work}, it is tasked with interpreting user inputs and determining appropriate actions to assist the user. When a user query is received, the LLM is provided with a system prompt containing instructions to process the query within a specific context. The system prompt designed to provide context to LLM which which enables the LLM to perform appropriate action in each steps.

\subsubsection{Text-Based Topological Map}
To generate our text-based topological map (Fig.~\ref{textmap+diagram}), we first construct an occupancy grid map of the area and then manually place nodes at each intersection, turn-able location, and critical navigation decision point. These nodes form the vertex set \(V\) of a directed and weighted graph \(\mathcal{G} = (V, E)\), where each edge \(e_{ij}\) connects node \(i\) to node \(j\) and is annotated with a distance \(d_{ij}\) and a relative direction \(\theta_{ij}\). The straight edges that connect these nodes not only reflect direct traversable paths but also encapsulate the metric and directional information necessary for navigation.
When embedded in a Euclidean plane, each node is assigned a coordinate \(p_i = (x_i, y_i)\), and the spatial transformation from node \(i\) to node \(j\) is expressed by
\[
p_j = p_i + \left( d_{ij}\cos\theta_{ij},\; d_{ij}\sin\theta_{ij} \right).
\]
We intentionally avoid curved paths or minor direction changes to maintain simplicity, in accordance with findings~\cite{shafique2024path,swobodzinski2009indoor} that suggest straight paths are more comfortable for individuals with visual impairments.

\subsubsection{Path Planning Module}
The planning module works in tandem with the agent to plan efficient routes to designated destinations. Upon a navigation query, it processes the text-based topological map, 
and generates a textual description of the route for the agent to process. This approach allows the agent to assess multiple paths and adapt in real-time based on hazards or user preferences.

\subsubsection{Embedding Module}
We use off-the-shelf CLIP \cite{clip} to generate embeddings from visual data, which are then stored in a vector database alongside location and orientation metadata. This metadata allows the LLM to reference the location and orientation of specific images upon retrieval. The embedding module is essential for localization and sub-goal selection, as it supports both text-to-image and image-to-image similarity searches.

\subsubsection{Vector Database}
The vector database stores high-dimensional representations of the environment generated by the embedding module, enabling real-time comparison of current observations with previously stored data. The agent accesses this data by querying navigation-related images and uses them to compare current observations with retrieved images. 

\subsubsection{Low-Level Planner}
The low-level planner converts the high-level decisions made by the agent into executable actions for the robot. It ensures that these commands are physically feasible, considering the robot's movement capabilities. 

\subsubsection{Robot}
We use TurtleBot as the robotic platform.

\subsection{Global Path Planning}
Figure~\ref{frame-work} illustrates the overall process of the proposed framework. We start with the assumption that the text map and images in the main vector database are pre-labeled. The process begins with the agent processing the user's query and the system prompt. The agent then generates high-level plans, navigational image query commands, and responses for the user.
As shown in Fig.~\ref{frame-work}, each output is communicated separately for clear communication between modules. High-level plans and user responses are delivered through a voice-to-text interface, while image queries retrieve navigation-related images from the vector database. These images are then embedded into a secondary vector database, which refines the navigation process by reducing ambiguity and ensuring that only relevant images are retrieved.

\subsection{Topological Navigation}
The agent begins navigation by querying the secondary vector database to retrieve the image corresponding to the current node. The retrieved image is compared to the current observation using cosine similarity. To improve place recognition, a similarity check is performed each time the robot travels to the next node or makes a turn. The low-level planner measures the distance traveled to the next node, and upon arrival, it sends a message to the agent. If the similarity score exceeds a predefined threshold, the agent concludes that the target node has been reached and generates the next set of movement commands. This process continues until the final destination is reached.

\subsection{Localization and Error Handling}
Localization in the Guide-LLM framework employs a dual-layered approach. Initially, the agent verifies its arrival at the desired node by querying the navigational vector database and comparing the retrieved image with the robot's current observation. If the similarity score between these images falls below a set threshold, the agent detects a potential localization error, suggesting that the robot may be in the wrong location. To address this, the agent initiates a broader search by querying the main vector database, which contains embeddings of the entire environment. This fallback mechanism allows Guide-LLM to re-localize by finding the most relevant match among all environment images. This dynamic error-handling process ensures that even if an agent is initially misaligned or disoriented,  Guide-LLM can correct its location in real-time, maintaining the desired path for navigation.

\subsection{Utilizing LLM’s Commonsense and Reasoning} 
Our framework utilizes the LLM's commonsense reasoning to improve navigation safety and decision-making. Unlike traditional systems that depend on predefined rules, the agent can interpret dynamic, real-world contexts to anticipate potential risks. For instance, if the agent identifies hazards such as a wet floor, warning tape, or unexpected obstacles through visual data or environmental descriptions, it proactively alerts the user and suggests an alternative route. The LLM's reasoning capabilities allow it to detect potential hazards even if they are not explicitly mentioned what is hazard. This flexibility enables the agent to adapt to changing conditions that rule-based systems might fail. By integrating commonsense knowledge, the agent enhances both the safety and overall reliability of the navigation experience.

\subsection{Personalization Potential}
A key strength of Guide-LLM is its ability to personalize the navigation experience according to each user’s specific preferences and needs. The agent's natural language interaction capabilities enable it to adjust its behavior in real-time, facilitating this personalization. For example, it can modify its path-planning to match a user's preferred walking speed, route preferences (e.g., avoiding stairs or choosing quieter areas), or specific safety concerns, such as steering clear of potential hazards. Additionally,  Guide-LLM can incorporate feedback from previous interactions, progressively refining its decisions to align with the user’s habits and preferences. For instance, if a user consistently opts for longer, less crowded routes over shorter ones, the agent can integrate this preference into its future planning. This adaptability extends beyond navigation; the agent can engage in personalized dialogue, adjusting the level of detail and communication style to meet different users' needs, whether they prefer brief instructions or more detailed explanations.


\begin{table}[t]
    \centering
    \renewcommand{\arraystretch}{1.3} 
    \caption{LLM Performance Comparison on Navigation (Success Rates)}
    \label{LLM_Performance}  
    \small 
    \begin{tabular}{p{5cm}cc} 
        \hline
        \textbf{Model (Parameters)}        & \textbf{House} & \textbf{Office} \\
        \hline
        GPT-4o~\cite{openai_gpt4}         & 84\%                            & 86\%                            \\
        Claude 3.5 Sonnet~\cite{claude3.5}& 74\%                            & 78\%                            \\
        Llama 3.2 Vision 90B~\cite{llama3.2}
                                          & 76\%                            & 74\%                            \\
        Llama 3.2 Vision 11B~\cite{llama3.2}
                                          & 20\%                            & 24\%                            \\
        Qwen2.5-VL-7B~\cite{qwen2.5-vl}   & 0\%                            & 4\%                            \\
        \hline
    \end{tabular}
\end{table}



\section{Results}
We evaluated the Guide-LLM by testing its ability to navigate through a simulated indoor environment based on user queries. The simulations are divided into scenarios designed to highlight distinct aspects of the Guide-LLM's utility in navigation, decision-making, error handling, and hazard detection and personalization. 

\subsection{Experiment Setup}
We used iGibson \cite{li2021igibson} simulator to validate Guide-LLM. The simulator was deployed using AWS G4dn.2xlarge instance and LLM models were deployed in the cloud using API.

\subsection{Model Selection and Topological Navigation}

To assess different LLM's ability to reach specified destinations and select optimal LLM for Guide-LLM, we performed evaluations in two indoor environments within the iGibson simulator [49]. These environments included a three-room house and a larger office-like setting with multiple hallways and offices, as shown in Fig. \ref{textmap+diagram}. In both environments, the agent needed to reach a specified destination using four key components: (1) a user query, (2) a text-based topological map, (3) a vector database, and (4) its own reasoning ability.
To examine the impact of model scale on navigation performance, we selected five vision-enabled LLMs: three models with over 90 billion parameters and two models with fewer than 11 billion parameters. We repeated each trial 50 times per environment to ensure statistical robustness. If the agent successfully arrived at the destination, we continued issuing new navigation queries until it eventually failed to reach a subsequent destination. If the agent failed to reach its current destination, we reset the robot’s position and counted that attempt as a failure.
Table \ref{LLM_Performance} shows that GPT-4o\cite{openai_gpt4}  achieved the highest overall success rate. Most navigation failures resulted from the LLM “forgetting” which command it should generate or confused with turning directions, suggesting that limited real-world grounding can hinder consistent decision-making. Furthermore, our experiments reveal that model size plays a crucial role in navigation success rates: GPT-4o’s performance was approximately three times higher than that of Llama 3.2 Vision 11B, suggesting a positive correlation between parameter count and navigation capability.

\begin{table}[t]
    \centering
    \caption{Localization Error Detection and Recovery}
    \label{localization}
    \renewcommand{\arraystretch}{1.3} 
    \begin{tabular}{p{4cm}c}  
        \hline 
        \textbf{Error Scenario}          & \textbf{Success Rate (\%)} \\ 
        \hline
        Localization Error Detection     & 90\% \\  
        Recovery                         & 72\% \\  
        \hline 
    \end{tabular}
\end{table}

\begin{table}[t]
\centering
\caption{Hazard Detection Results}
\label{tab:hazard_results}
\begin{tabular}{lcccccc}
\hline
\textbf{Scenario} & 
\textbf{\# Trials} &
\textbf{TP} &
\textbf{FP} &
\textbf{FN} &
\textbf{TN} \\
\hline
Potential Hazard     & 30 &  25  &     &  5  &   \\
Non-Potential Hazard & 30 &   &  10  &   &  20  \\
\hline
\end{tabular}
\end{table}


\subsection{Localization Error Detection and Recovery}
To evaluate Guide-LLM's resilience against localization failures, we introduced the `kidnapped robot' scenario in the office environment. During navigation, the robot was randomly teleported to a different node, forcing the model to detect error and recover from it. We performed this evaluation 50 times in total, ignoring naturally failed scenarios where navigation failed due to LLM's navigation failure.
If the agent successfully recognized the mismatch between its expected location and current observation, we considered the localization error detection successful. 
If the agent re-localize and reached the goal by updating its plan based on the new location, the trial was marked as a successful recovery. Table~\ref{localization} shows the localization error detection and recovery rates. 
In addition, we observed that most failures occurred in feature-sparse corridors or visually similar places. We also observe that recovery failures happen when the Guide-LLM recognized the localization error but carries on with the previous path.

\subsection{Commonsense Reasoning for Hazard Detection}
The objective of this experiment was to evaluate the agent’s ability to identify potential hazards in the environment, communicate these risks to the user, and adapt its actions based on user decisions. To test this capability, we placed obstacles and potential hazards (warning signs, physical barriers) and non-potential hazards along the navigation path (chair, pot). As shown in Table~\ref{tab:hazard_results}, we conducted 60 trials: 
30 with objects and hazards with the potential to impede navigation, 
while the other 30 with non-hazardous objects that did not interfere with safe travel. In the hazard scenario, we recorded 25 true positives (TP), where the agent correctly identified a hazard and warned the user, and 5 false negatives (FN), where it failed to recognize a hazard and proceeded without warning. These false negatives typically resulted from limited height awareness of both the Guide-LLM and the user.
In the non-hazard scenario, we observed 10 false positives (FP), where the agent mistakenly flagged hazard, and 20 true negatives (TN), where it ignored a harmless item.
Figure~\ref{textmap+diagram} illustrates a typical hazard scenario in which the agent demonstrates commonsense reasoning beyond navigation. Upon detecting a hazard, it alerts the user and proposes an alternate route, underscoring the system’s capacity to dynamically adapt when faced with unexpected obstacles.

\subsection{Personalization Potential}
Prior research on navigational aids indicates that personalization is vital for enhancing user experience~\cite{ahmetovic2019impact}. As shown in Table~\ref{tab:improved_comparison}, current navigation aids are limited in their ability to incorporate personalized preferences. In contrast, Guide-LLM supports diverse user needs through natural language interaction, enabling users to easily communicate specific requirements. To illustrate how user preferences can be addressed, we renamed certain spaces in the text-based topological map (e.g., “concert hall,”, "food court", “noisy area,” “quiet area”). We then conducted 50 repeated trials on the same route, applying various user preferences and phrasings (e.g., avoiding stairs, noisy areas, or avoiding places likely to be crowded). In each trial, Guide-LLM adapted its path planning accordingly, even selecting routes up to three times longer if it aligned with the user’s stated requirements. This is possible by the integration of the text-based map with the path planning module, allowing the Guide-LLM to evaluate alternative routes and plan ahead. Some failures were seen when the number of possible routes was excessively large, potentially exceeding the maximum context window and leading to suboptimal decisions. This test resulted 92\% success rate. Additionally, we evaluated Guide-LLM’s ability to adjust speed by providing multiple instructional cues (e.g., asking it to speed up or slow down) to see whether the system could infer user intent. We observed that Guide-LLM consistently increased or decreased walking speed without any failures.
For characteristic changes such as providing detailed, step-by-step navigation or shifting conversational tone (e.g., formal, casual, or informal), the system met user expectations in 43 out of 50 trials.

\subsection{LLM Guided Navigation Ablation Study}
We conducted an ablation study to examine how each core component of the Guide-LLM framework contributes to navigation performance. As shown in Table~\ref{LLM_Performance}, the full system achieved an 86\% success rate when navigating through an office environment. We then systematically removed individual components and measured the resulting performance drops.
Removing the system prompt reduced the success rate to 0\%, indicating that prompt-driven context is essential for the agent’s high-level understanding and instruction formulation.
Next, we removed the path planning module and directly supplied the text-based map to the agent. This change lowered the success rate to 40\%. The primary reason is that adding a large text map to the system prompt can exceed the context window, which degrades the agent's reasoning capability due to processing bottlenecks associated with large prompts ~\cite{an2023eval,bai2023longbench,levy2024same}.
Finally, removing the image retrieval system forced users to specify the starting location, thereby altering the experimental conditions and increasing user burden. Since it no longer tested the agent's ability to visually infer starting points, we excluded this configuration from our comparative analysis to maintain fairness.

\section{Conclusion and Future Work}
We introduce Guide-LLM, an embodied large language model framework that can provide efficient, adaptive, and personalized navigation assistance for people with visual impairments. 
By combining a text-based topological map with common sense reasoning, our system  reduces the need for explicit user instructions, enabling personalized navigation, offering a more intuitive navigational experience.
Simulation results demonstrate that Guide-LLM excels in tasks such as localization error detection recovery, hazard detection, and personalization. 
As this research focused on developing and validating the technical feasibility of LLM based navigation aid, we plan to focus on the following directions in the future: Optimizing smaller language models to match or approximate the performance of larger ones, allowing on-device processing without reliance on cloud-based services. Expanding the text-based map creation process to autonomously generate and update environmental representations in large, complex buildings such as multi-story shopping center.
Attaching depth camera to provide depth information to improve on hazard detection. Conducting real-world trials with visually impaired participants to evaluate usability, comfort, and overall effectiveness in navigation tasks. 
By addressing these challenges, we aim to refine Guide-LLM into a comprehensive assistive solution that empowers visually impaired individuals with more flexible, accessible, and safe navigation.

\addtolength{\textheight}{-4cm}   







\newpage
\bibliographystyle{IEEEtran}  
\bibliography{references}  

\end{document}